\documentclass[conf]{new-aiaa}
\usepackage[utf8]{inputenc}

\usepackage{graphicx}
\usepackage{amsmath}
\usepackage[version=4]{mhchem}
\usepackage{siunitx}
\usepackage{longtable,tabularx,subfigure,wrapfig}

\usepackage{makecell}
\usepackage{caption}
\usepackage{subcaption}
\usepackage{wrapfig}
\usepackage{siunitx}
\usepackage{multirow}

\usepackage{tabularray}
\usepackage{enumitem}
\usepackage{comment}

\newcolumntype{C}[1]{>{\centering\arraybackslash}p{#1}}

\title{The Use of Multi-Scale Fiducial Markers To Aid Takeoff and Landing Navigation by Rotorcraft	}

\author{Jongwon Lee\footnote{Ph.D. Student, Aerospace Engineering, Email: jongwon5@illinois.edu}, Su Yeon Choi\footnote{Ph.D. Student, Aerospace Engineering, Email: suyeonc3@illinois.edu}, and Timothy Bretl\footnote{Professor, Aerospace Engineering, Email: tbretl@illinois.edu}}
\affil{University of Illinois at Urbana-Champaign, Urbana, Illinois, 61801}

\begin{document}

\maketitle

\begin{abstract}

This paper quantifies the performance of visual SLAM that leverages multi-scale
fiducial markers (i.e., artificial landmarks that can be detected at a wide range of distances)
to show its potential for reliable takeoff and landing navigation in rotorcraft. Prior work has
shown that square markers with a black-and-white pattern of grid cells can be used to improve
the performance of visual SLAM with color cameras. We extend this prior work to allow nested marker layouts. We evaluate performance during semi-autonomous takeoff and landing operations in a variety of environmental conditions by a DJI Matrice 300 RTK rotorcraft with two FLIR Blackfly color cameras, using RTK GNSS to obtain ground truth pose estimates. Performance measures include absolute trajectory error and the fraction of the number of estimated poses to the total frame. We release all of our results --- our dataset and the code of the implementation of the visual SLAM with fiducial markers --- to the public as open-source.

\end{abstract}

\section{Introduction}

\lettrine{V}{isual} SLAM with fiducial markers, a variation of simultaneous localization and mapping (SLAM), utilizes easily detectable and identifiable artificial visual patterns called fiducial markers to aid in mapping and tracking. Several previous studies~\cite{lim2009real,yamada2009study,pfrommer2019tagslam,munoz2020ucoslam,lee2023comparative} have shown that visual SLAM with fiducial markers offers improved performance compared to generic visual SLAM, which may enhance the reliability of navigation scenarios during takeoff and landing, adhering to visual flight rules.

While visual SLAM approaches with fiducial markers offer advantages, the existing literature comes with potential ambiguity when applied to the navigation of rotorcraft during takeoff and landing.
The primary concern arises from the common practice of using fiducial markers of uniform size in existing works, which constrains their detectable distance range. This limitation may impact SLAM performance significantly, especially when applied to takeoff and landing scenarios, where the distance between markers on the ground and the camera on the rotorcraft varies significantly.
Moreover, existing visual SLAM approaches with fiducial markers are primarily assessed only in indoor environments, where visibility conditions remain constant. It is crucial to investigate how such visual SLAM approaches perform under various outdoor visibility conditions, including different illumination levels and adverse weather, which are likely to be encountered during actual takeoff and landing navigation in rotorcraft.

\begin{figure}[hbt!]
  \centering
  \begin{minipage}[t]{0.4\textwidth}
    \centering
    \includegraphics[width=\textwidth]{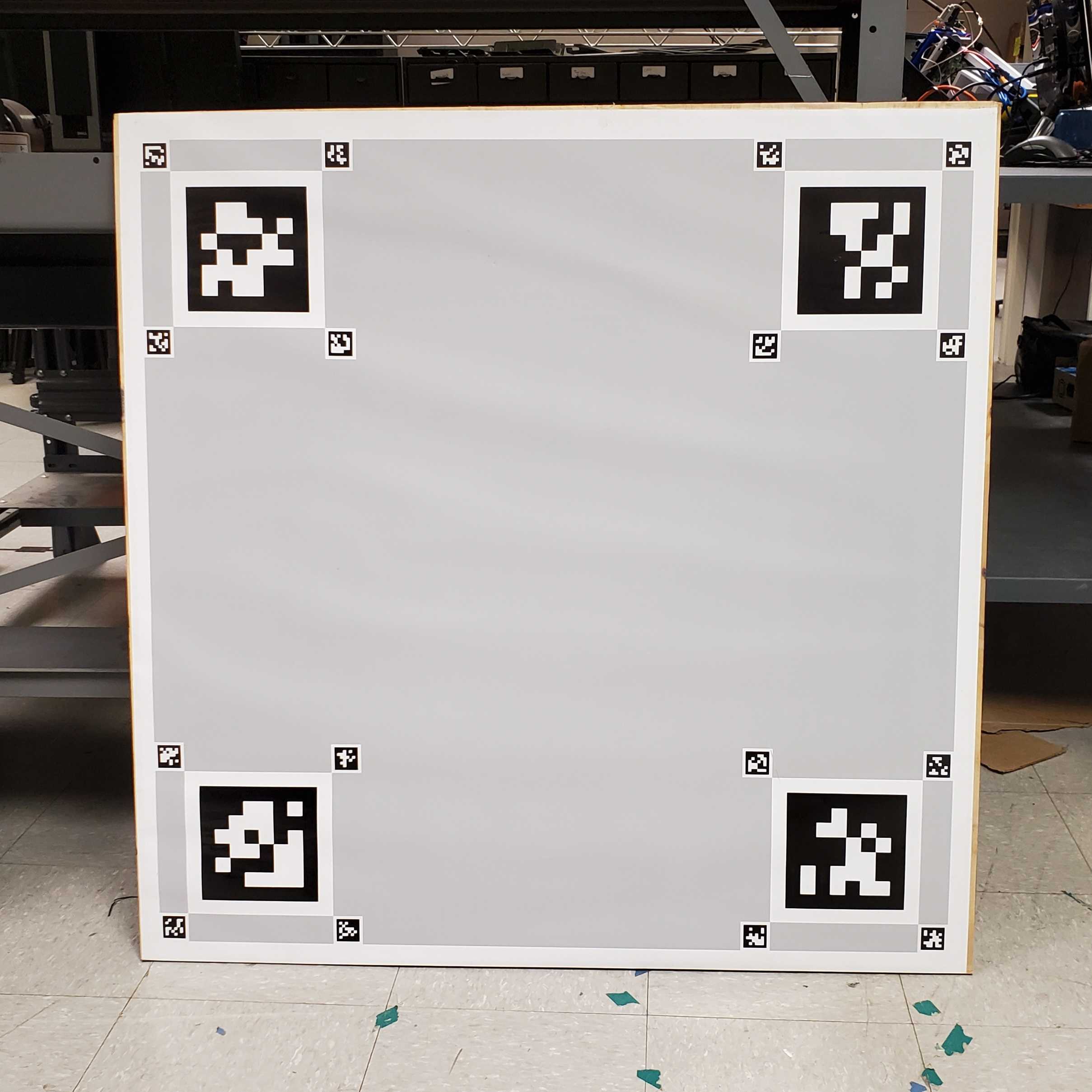}
  \end{minipage}
  \hspace{1.0cm}
  \begin{minipage}[t]{0.4\textwidth}
    \centering
    \includegraphics[width=\textwidth]{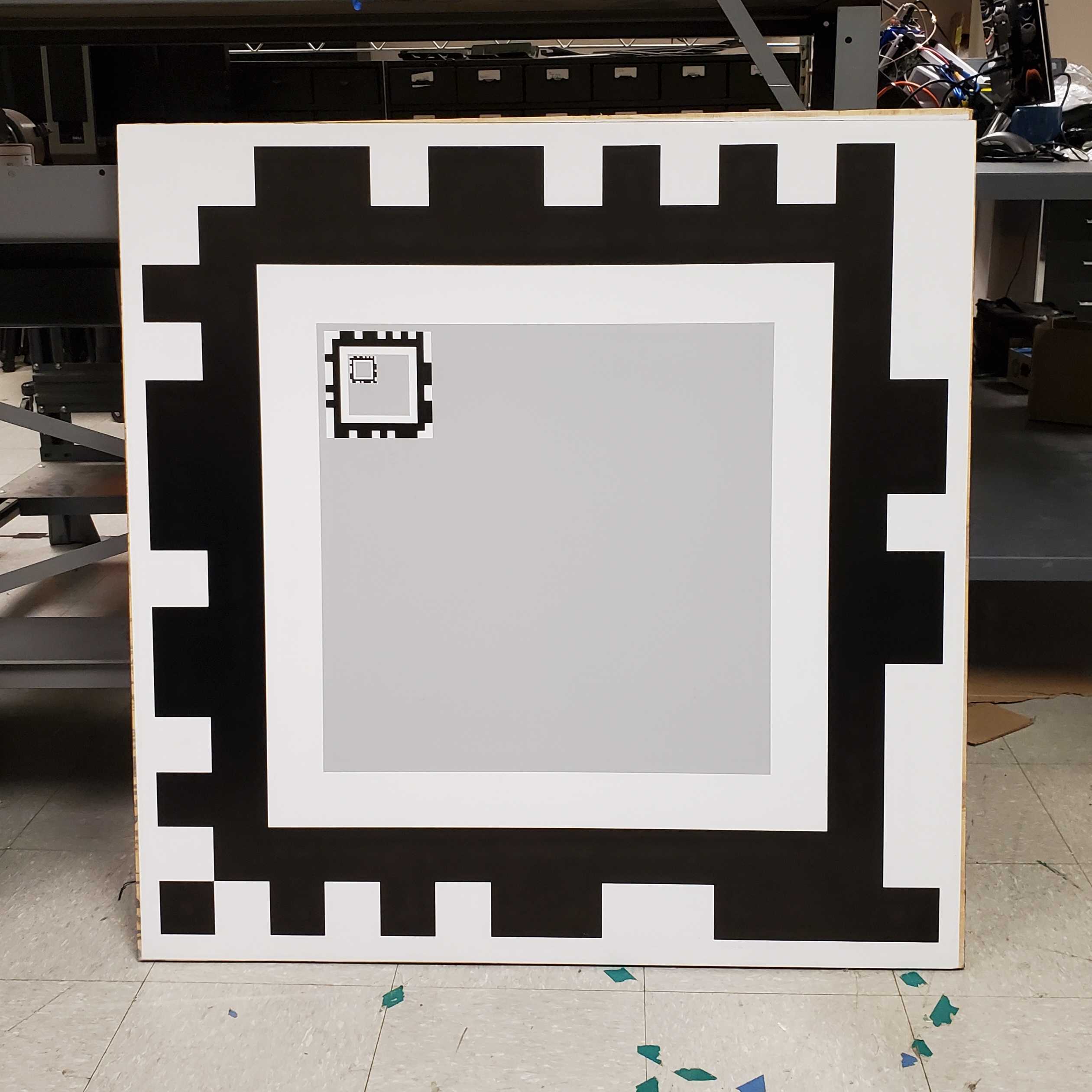}
    \captionsetup{justification=centering} 
  \end{minipage}
  
  \caption{Multi-scale fiducial markers proposed for use. Non-nested layout with AprilTag Standard36h11 family (left) and nested layout with AprilTag Custom52h12 family (right) integrated into the touchdown and liftoff area (TLOF), adhering to FAA guidelines for vertiport design~\cite{faa}.}
  \label{fig: Multi-scale fiducial markers}
\end{figure}

In response to these limitations, our investigation focuses on two key contributions within the realm of visual SLAM with fiducial markers.
Firstly, we introduce the utilization of multi-scale fiducial markers, derived from a set with flexible layouts~\cite{krogius2019apriltag}, showcased in Fig.~\ref{fig: Multi-scale fiducial markers}. This approach enables detection across a wider range of distances, addressing the limitations highlighted in a prior work proposing the use of fiducial markers for rotorcraft navigation~\cite{springer2022autonomous}. 
Secondly, we assess the performance of visual SLAM with multi-scale fiducial markers on a dataset collected outdoors with a rotorcraft. This dataset emulates semi-autonomous takeoff and landing operations performed by a DJI Matrice 300 RTK rotorcraft in various environmental conditions. The dataset includes image data captured by two FLIR Blackfly color cameras, with ground truth pose estimates obtained using RTK GNSS.

Section~\ref{Section: Related Works} delves into various related works, with a specific focus on introducing the concept of visual SLAM with fiducial markers. The subsequent section, Section~\ref{Section: Experiments}, outlines the system we devised for collecting data in semi-autonomous takeoff and landing scenarios governed by visual flight rules. This section also covers details about the employed multi-scale fiducial marker on the vertiport, the flight scenario implemented, and the SLAM code utilized. The evaluations and discussions are presented comprehensively in Section~\ref{Section: Results} and \ref{Section: Discussion}, respectively, and the paper concludes with a summary and remarks in Section~\ref{Section: Conclusion}.

Both the code and dataset used in this paper are available online\footnote{\url{https://github.com/tag-nav/wolf_ros}}.

\section{Related Works}
\label{Section: Related Works}

SLAM is a process through which a mobile robot constructs a map of its environment while simultaneously determining its own location within that map. A specific subset of SLAM utilizing image data from one or more cameras is known as visual SLAM. In visual SLAM, the process typically involves extracting features from the current image, associating them with features from previous images, and concurrently estimating the poses of landmarks (map) and the camera's trajectory.

The use of square markers featuring a black-and-white grid pattern, commonly known as fiducial markers, has gained widespread adoption in robotics applications. These markers serve as easily identifiable visual landmarks with a low probability of misdetection. While some works~\cite{lim2009real,yamada2009study,pfrommer2019tagslam} solely rely on the detection results of these fiducial markers --- rather than utilizing feature points like corners, a commonly employed landmark information --- UcoSLAM~\cite{munoz2020ucoslam}, an approach based on a feature point based state-of-the-art visual SLAM approach~\cite{mur2017orb}, proposes the simultaneous use of marker detection results and feature points. This hybrid approach shows enhanced performance compared to solutions relying solely on either fiducial markers or feature points alone.

While the method UcoSLAM introduces --- the SLAM approach using both marker and feature point detection results --- positions itself as a viable choice for visual SLAM with fiducial markers, its implementation\footnote{\url{https://www.uco.es/investiga/grupos/ava/portfolio/ucoslam/}} has a few limitations. Firstly, the UcoSLAM implementation is tied to a specific type of fiducial marker known as ArUco markers~\cite{garrido2014arucomarker}, precluding the use of Apriltag~\cite{krogius2019apriltag} with flexible layouts, which allows for the utilization of multi-scale fiducial markers. Secondly, the UcoSLAM implementation poses challenges when it comes to extending its functionality to incorporate other types of sensor measurements typically found on a mobile robot, such as IMU and GNSS. This limitation hinders its future potential extensions. In contrast, WOLF~\cite{sola2022wolf}, an open-source modular SLAM framework, overcomes these constraints. It offers a visual SLAM implementation with Apriltag~\cite{krogius2019apriltag} --- again, the type of widely used fiducial marker allowing flexible layouts for designing multi-scale fiducial markers ---- and is easily extendable to various sensor configurations, providing the potential for diverse extensions in future development.

\section{Experiments}
\label{Section: Experiments}

\subsection{System for data collection}

\begin{figure}[hbt!]
    \centering
    \includegraphics[width=0.80\textwidth]{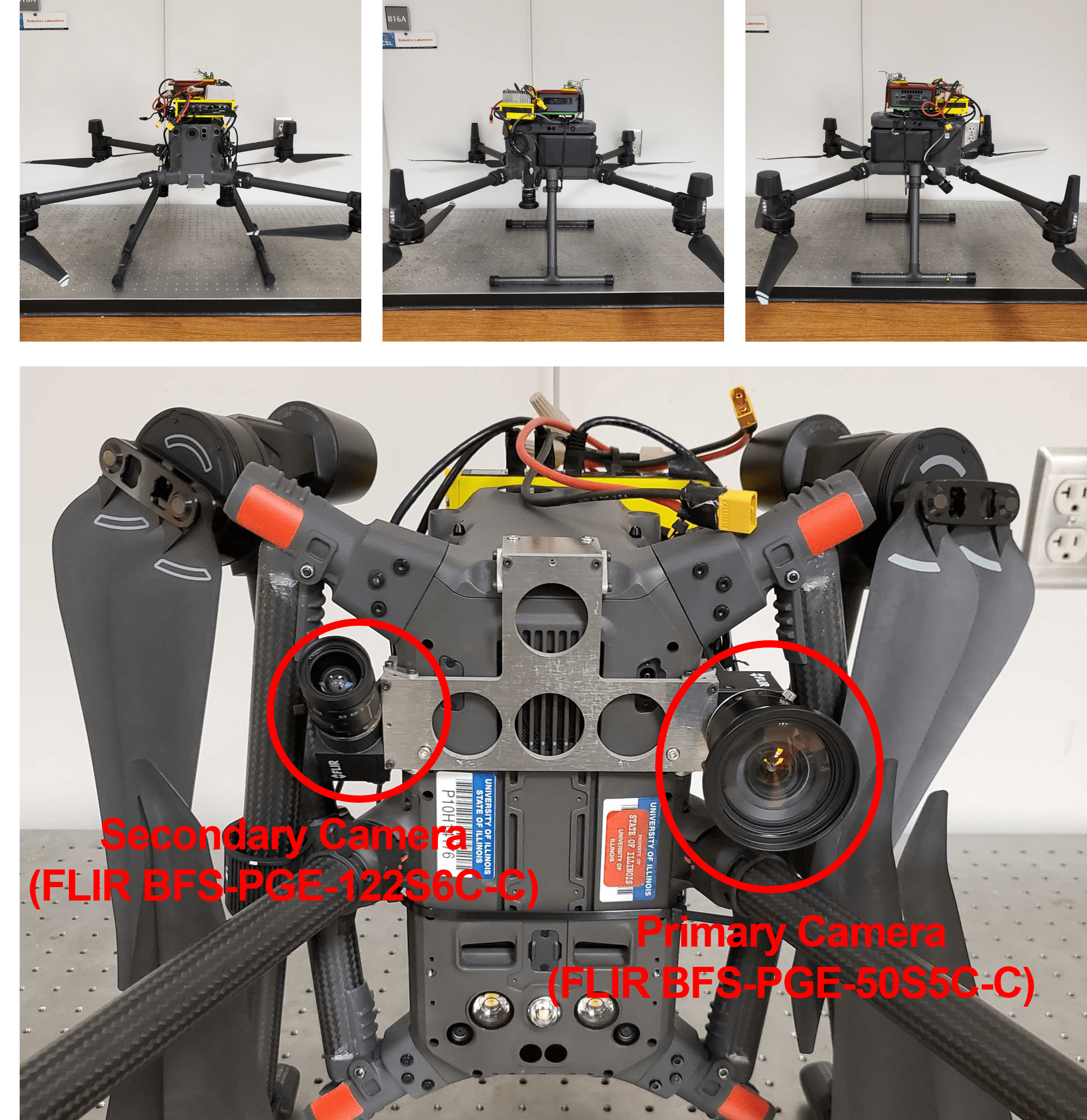}
    \captionsetup{justification=centering} 
    \caption{Front and side views of the DJI Matrice 300 RTK rotorcraft equipped with a sensor system for data collection (top). The bottom of the rotorcraft hosts a sensor system comprising two cameras, one directed downward and the other positioned at a 45$^{\circ}$ forward angle (bottom).}
    \label{fig: Sensor system}
\end{figure}

Fig.~\ref{fig: Sensor system} illustrates the DJI Matrice 300 RTK rotorcraft utilized for our data collection. The rotorcraft is equipped with RTK GNSS capabilities, offering enhanced precision in measurements compared to standard GNSS. This capability is crucial for providing accurate ground truth data in the evaluation of SLAM. Two FLIR Blackfly color cameras are mounted at the bottom --- one BFS-PGE-50S5C-C with a resolution of 2448x2048 facing downward (primary) and the other BFS-PGE-122S6C-C with a resolution of 4096x3000 oriented 45$^{\circ}$ forward (secondary). This configuration is designed to capture image data during flight, with only a slight overlap between the cameras, ensuring a broad field of view for the easy detection of fiducial markers on the ground.

\subsection{Dataset}

\subsubsection{Multi-scale fiducial markers}

We propose the utilization of two types of multi-scale fiducial markers, namely non-nested and nested layouts, depicted in Fig.~\ref{fig: Multi-scale fiducial markers}. These layouts are based on the Standard36h11 and Custom52h12 Apriltag~\cite{krogius2019apriltag} families. The rationale behind incorporating fiducial markers at multiple scales is to extend the detectable distance range. For example, experimental findings presented by \citeauthor{krogius2019apriltag}~\cite{krogius2019apriltag} indicate that a fiducial marker with a unit side length can be consistently detected from distances ranging from 5 to 20 units. Moreover, other prior works~\cite{springer2022autonomous,kalaitzakis2021fiducial} underscore the restricted detectable distance range of single-scale fiducial markers. This emphasizes the need for employing multi-scale markers to extend the range, a capability not attainable with single-scale markers. Consequently, employing markers of various sizes enhances the robustness of visual SLAM systems, ensuring more reliable performance compared to using markers of a single size.

Returning to the specifics, the non-nested layout integrated into the touchdown and liftoff area (TLOF), following FAA guidelines for vertiport design~\cite{faa}, consists of twenty Standard36h11 Apriltag markers with three different scales (1:5:28). The nested layout integrated into the vertiport comprises three Custom52h12 Apriltag markers with three different scales (1:4:30). These markers are printed in a 1m$^2$ size to align with the control dimension of the DJI Matrice 300 RTK, the rotorcraft used for data collection.

\subsubsection{Data collection under the scenario encompassing takeoff and landing of rotorcraft}

\begin{figure}[hbt!]
    \centering
    \includegraphics[width=0.6\textwidth]{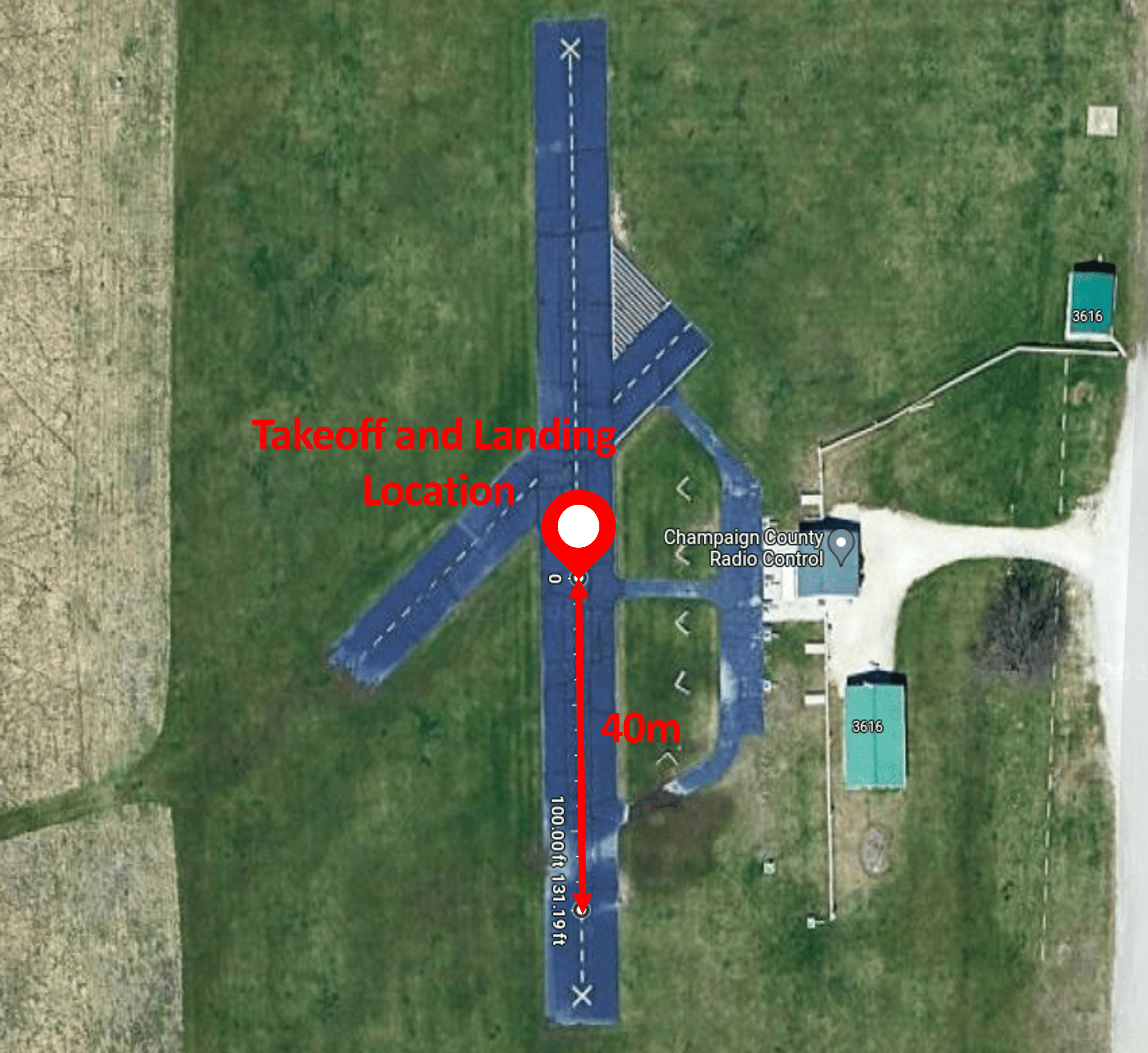}
    \captionsetup{justification=centering} 
    \caption{Champaign County R/C field overlaid with a marker indicating takeoff and landing location, along with the depicted flight trajectory.}
    \label{fig: Trajectory profile}
\end{figure}

\begin{figure}[hbt!]
    \centering
    \includegraphics[width=0.9\textwidth]{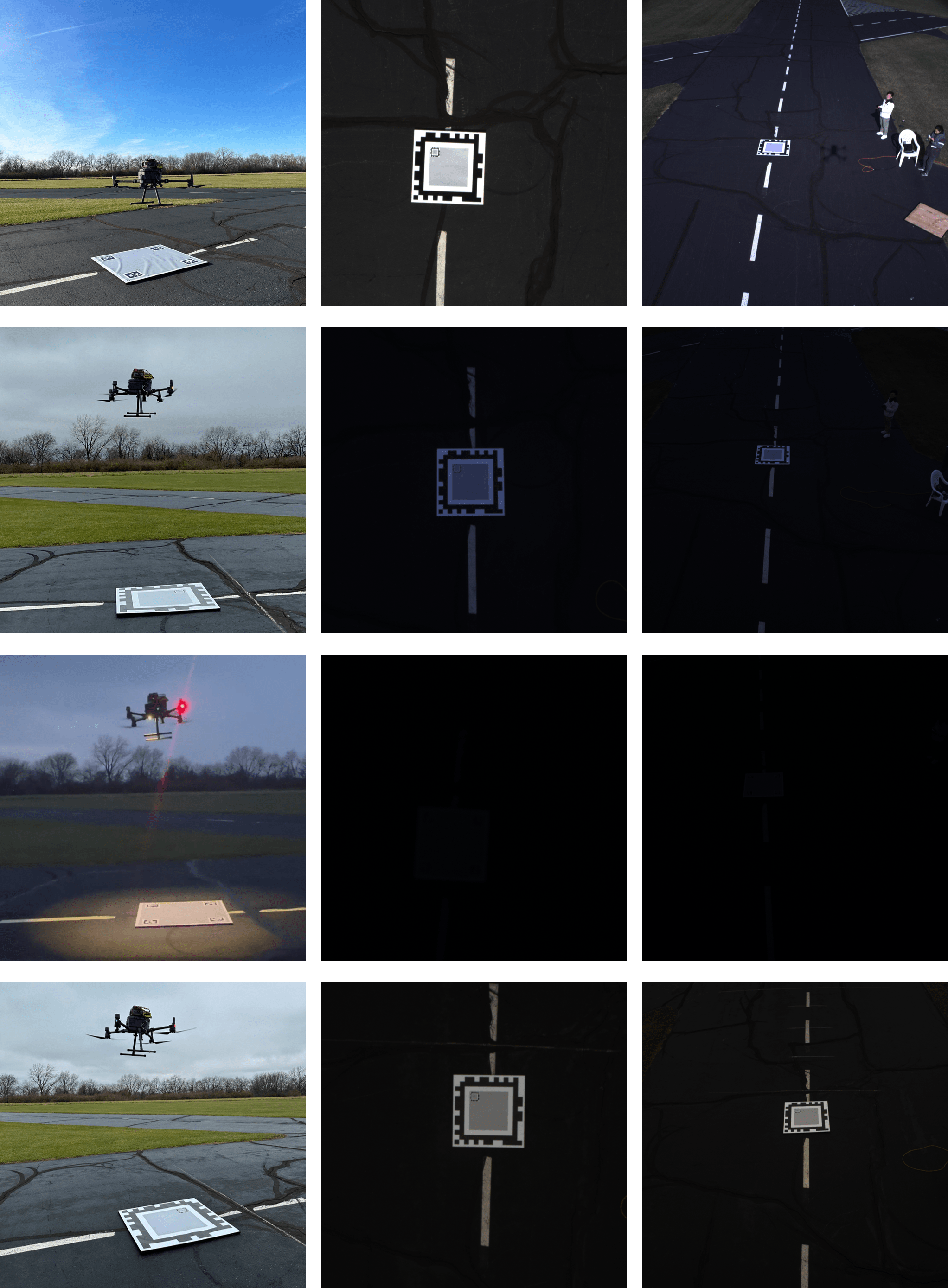}
    \captionsetup{justification=centering} 
    \caption{Rotorcraft flights conducted for data collection under diverse weather conditions (left column), accompanied by examples of images captured from the primary camera (mid column) and the secondary camera (right column) during each flight mission.}
    \label{fig: Experiments}
\end{figure}

We implement a trajectory encompassing both the takeoff and landing phases of the rotorcraft. Initially, the rotorcraft ascends vertically to an altitude of 5 meters above ground level (AGL). Subsequently, it traverses a horizontal distance of 40 meters at a speed of 1 m/s. After a pause, the rotorcraft returns to a location directly above the vertiport at a 5-meter altitude, followed by the landing phase. These maneuvers adhere to visual flight rules (VFR) approach/departure path requirements, maintaining an 8:1 ratio in horizontal and vertical units~\cite{faa}. Throughout the entire flight, a human operator remotely controls the rotorcraft.

The dataset was collected at Champaign County R/C field (3616 W Bloomington Rd, Champaign, IL 61822, USA) as shown in Fig.~\ref{fig: Trajectory profile}, providing a suitable location for rotorcraft flights. Multi-scale fiducial markers, as described earlier, were positioned along the 400-foot-length runway. The rotorcraft followed the specified trajectory along the southbound runway during both takeoff and landing simulations. Data collection occurred on two distinct dates, November 30th, 2023, and December 2nd, 2023, capturing various times and weather conditions to encompass different visibility scenarios as shown in Fig.~\ref{fig: Experiments}.

\subsection{SLAM implementation}

We employed WOLF~\cite{sola2022wolf}, an open-source modular SLAM framework that already incorporates a visual SLAM implementation with Apriltag~\cite{krogius2019apriltag}. Specifically, we set up a binocular visual SLAM system using the two types of Apriltag-based multi-scale fiducial markers we proposed earlier --- nested and non-nested layouts as shown in Fig.~\ref{fig: Multi-scale fiducial markers} --- by the two synchronized FLIR Blackfly color cameras. The intrinsic and extrinsic parameters of these cameras were identified using Kalibr~\cite{furgale2013unified}, an open-source camera calibration tool.

In what follows, we evaluate the two configurations of visual SLAM with fiducial markers provided by WOLF~\cite{sola2022wolf}. The first mode relies solely on marker detection results to construct visual landmarks (marker SLAM), while the second mode utilizes both marker and feature detection results for landmark construction (marker + feature SLAM). This aims to investigate which mode performs well in our rotorcraft takeoff and landing scenarios under visual flight rules in diverse conditions.

\section{Results}
\label{Section: Results}

Tables~\ref{table: non-nested results} and \ref{table: nested results} show the results for visual SLAM using non-nested and nested multi-scale fiducial markers, depicted in the left and right images of Fig.~\ref{fig: Multi-scale fiducial markers}, respectively, under various conditions. The evaluation metrics include the absolute trajectory error (ATE; lower is better) and the fraction of the number of estimated poses to the total frame, which represents the percentage of time the navigation system is operational and usable by the aircraft (availability; higher is better). ATE is a standard performance-based requirement for navigation systems utilizing SLAM in the robotics community. The availability measurement aligns with the performance requirements outlined by ICAO for performance-based navigation.

We also assessed the performance of ORB-SLAM3~\cite{campos2021orb}, a state-of-the-art visual SLAM utilizing feature points, on our dataset. However, it consistently failed in all cases, and therefore, we have excluded the results from the tables.

\begin{table}[hbt!]
\centering
\caption{Results for visual SLAM with non-nested multi-scale fiducial marker. Performance measures include absolute trajectory error (ATE) and the fraction of the number of estimated poses to the total frame (availability).}
\begin{tabular}{c|c|c|c|c|c|C{4.2em}|C{4.2em}|C{4.2em}|C{4.2em}}
\hline\hline
\multirow{2}{*}{Date} & \multicolumn{4}{c|}{Weather} & \multirow{2}{*}{Trial} & \multicolumn{2}{c|}{Marker SLAM} & \multicolumn{2}{c}{Marker + Feature SLAM} \\
\cline{2-5} \cline{7-10} 
 & state & temp. & wind & illumination & & ATE (m) & Availability & ATE (m) & Availability \\
\hline
\multirow{8}{*}{\shortstack{Nov. 30, \\ 2023}} & \multirow{3}{*}{sunny} & \multirow{3}{*}{\SI{10}\degreeCelsius} & \multirow{3}{3em}{5.3 m/s NE} & \multirow{3}{4.2em}{6000 Lux (day)} & 1 & 0.47 & 0.84 & 0.39 & 0.84 \\
 & & & & & 2 & 2.04 & 0.84 & 2.56 & 0.83 \\
 & & & & & 3 & 1.46 & 0.84 & 1.70 & 0.84 \\
\cline{2-10}
 & \multirow{3}{*}{drizzle} & \multirow{3}{*}{\SI{10}\degreeCelsius} & \multirow{3}{3em}{6.5 m/s NE} & \multirow{3}{4.2em}{1200 Lux (day)} & 1 & 2.19 & 0.84 & 2.58 & 0.84 \\
 & & & & & 2 & 3.56 & 0.84 & 4.92 & 0.84 \\
 & & & & & 3 & 1.86 & 0.85 & 1.59 & 0.84 \\
\cline{2-10}
 & \multirow{2}{*}{drizzle} & \multirow{2}{*}{\SI{9}\degreeCelsius} & \multirow{2}{3em}{7.3 m/s N} & \multirow{2}{4.2em}{10-50 Lux (dusk)} & 1 & - & - & - & - \\
 & & & & & 2 & - & - & - & - \\
\hline
 \multirow{4}{*}{\shortstack{Dec. 2, \\ 2023}} & \multirow{4}{*}{cloudy} & \multirow{4}{*}{\SI{6}\degreeCelsius} & \multirow{4}{3em}{1.3 m/s S} & \multirow{4}{4.2em}{4000 Lux (day)} & 1 & 2.00 & 0.84 & 1.82 & 0.84 \\
 & & & & & 2 & - & - & 4.95 & 0.84 \\
 & & & & & 3 & 0.33 & 0.84 & 0.54 & 0.84 \\
 & & & & & 4 & 0.49 & 0.89 & 1.61 & 0.83 \\
\hline\hline
\end{tabular}
\label{table: non-nested results}
\end{table}

\begin{table}[hbt!]
\centering
\caption{Results for visual SLAM with nested multi-scale fiducial marker. Performance measures include absolute trajectory error (ATE) and the fraction of the number of estimated poses to the total frame (availability).}
\begin{tabular}{c|c|c|c|c|c|C{4.2em}|C{4.2em}|C{4.2em}|C{4.2em}}
\hline\hline
\multirow{2}{*}{Date} & \multicolumn{4}{c|}{Weather} & \multirow{2}{*}{Trial} & \multicolumn{2}{c|}{Marker SLAM} & \multicolumn{2}{c}{Marker + Feature SLAM} \\
\cline{2-5} \cline{7-10} 
 & state & temp. & wind & illumination & & ATE (m) & Availability & ATE (m) & Availability \\
\hline
\multirow{8}{*}{\shortstack{Nov. 30, \\ 2023}} & \multirow{3}{*}{sunny} & \multirow{3}{*}{\SI{10}\degreeCelsius} & \multirow{3}{3em}{5.3 m/s NE} & \multirow{3}{4.2em}{6000 Lux (day)} & 1 & 1.00 & 0.80 & 0.77 & 0.82 \\
 & & & & & 2 & 0.92 & 0.80 & 1.11 & 0.81 \\
 & & & & & 3 & 0.69 & 0.80 & 0.78 & 0.81 \\
\cline{2-10}
 & \multirow{2}{*}{drizzle} & \multirow{2}{*}{\SI{10}\degreeCelsius} & \multirow{2}{3em}{6.5 m/s NE} & \multirow{2}{4.2em}{1200 Lux (day)} & 1 & - & - & - & - \\
 & & & & & 2 & 0.90 & 0.81 & 0.96 & 0.83 \\
\cline{2-10}
 & \multirow{3}{*}{drizzle} & \multirow{3}{*}{\SI{9}\degreeCelsius} & \multirow{3}{3em}{7.3 m/s N} & \multirow{3}{4.2em}{10-50 Lux (dusk)} & 1 & - & - & - & - \\
 & & & & & 2 & - & - & - & - \\
 & & & & & 3 & - & - & - & - \\
\hline
 \multirow{6}{*}{\shortstack{Dec. 2, \\ 2023}} & \multirow{6}{*}{cloudy} & \multirow{6}{*}{\SI{6}\degreeCelsius} & \multirow{6}{3em}{1.3 m/s S} & \multirow{6}{4.2em}{4000 Lux (day)} & 1 & 0.69 & 0.84 & 0.83 & 0.84 \\
 & & & & & 2 & 1.46 & 0.80 & 0.98 & 0.82 \\
 & & & & & 3 & 0.99 & 0.80 & 1.05 & 0.82 \\
 & & & & & 4 & 0.77 & 0.80 & 0.90 & 0.81 \\
 & & & & & 5 & 1.31 & 0.80 & 1.39 & 0.82 \\
 & & & & & 6 & 0.79 & 0.78 & 0.88 & 0.97 \\
\hline\hline
\end{tabular}
\label{table: nested results}
\end{table}

\section{Discussion}
\label{Section: Discussion}

One significant observation from the results presented in Section~\ref{Section: Results} is the failure of SLAM in all data collected under the lowest illumination condition (10-50 Lux). This failure is consistent across both types of multi-scale fiducial markers and whether using a marker SLAM or a marker + feature SLAM approach. The challenge in such low-light environments is attributed to the reduced visibility of fiducial markers, making detection challenging.

In comparing marker SLAM and marker + feature SLAM, no significant differences are evident in terms of both ATE and availability. This finding contradicts the argument presented by UcoSLAM~\cite{munoz2020ucoslam}, which advocates for the enhanced performance of using both marker and feature point detection results. The discrepancy may stem from our testing environment in which the rotorcraft flew over a runway with limited texture, hindering the effectiveness of feature point detection. The similarity in availability measures between marker and marker + feature SLAM modes supports this hypothesis, suggesting that not only the marker SLAM mode but also the marker + feature SLAM mode struggle to locate the rotorcraft once the fiducial marker on the vertiport is out of the field of view. This occurs as the rotorcraft moves farther away during the flight mission, as illustrated in Fig.~\ref{fig: Trajectory profile}.

We note that RTK GNSS measurements, while used as ground truth, may not be as accurate as expected. Significantly, we observed a notable mismatch in the takeoff and landing positions, which were intended to coincide, when examining the plotted RTK GNSS measurements across various missions. Consequently, readers are advised to interpret ATEs, which are evaluated against RTK GNSS measurements as ground truth, as indicative of visual SLAM with fiducial markers providing pose estimates within a certain boundary relative to RTK GNSS measurements. This recommendation is made instead of drawing rigorous conclusions about the superiority of one mode over another or its performance in specific conditions.

\section{Conclusion}
\label{Section: Conclusion}

This paper introduces the application of visual SLAM with multi-scale fiducial markers and assesses its performance using visual data captured by a pair of cameras in rotorcraft takeoff and landing scenarios across diverse weather conditions. Evaluation focuses on two metrics: absolute trajectory error and the fraction of estimated poses to the total frame. 

We recognize the potential benefits of integrating additional measurements, such as inertial data and GNSS, to enhance SLAM accuracy and efficiency. Future work particularly involves incorporating inertial measurement data from one or more IMUs~\cite{lee2022extrinsic} into the SLAM approach. Additionally, we acknowledge the opportunity to enhance accuracy by leveraging the known configurations of co-planar multi-scale fiducial markers through the adoption of a perspective-n-points (PnP) algorithm. Our next step includes integrating a PnP algorithm tailored to our multi-scale marker layouts and reporting its performance within the existing SLAM framework we use.

\section*{Acknowledgments}
This work is supported by Supernal, LLC.

\bibliography{reference}

\end{document}